\documentclass{article}
\usepackage{spconf,amsmath,epsfig}
\usepackage[center]{caption}
\usepackage{epstopdf}
\usepackage{array}
\usepackage{multicol}
\usepackage{threeparttable}
\usepackage{longtable}
\usepackage{fancyvrb}
\usepackage{changepage}
\usepackage[stable]{footmisc}
\usepackage[T1]{fontenc}
\usepackage{multirow}
\usepackage{amsmath}
\usepackage{amsfonts}
\usepackage{subfigure}
\usepackage{xcolor}
 \usepackage{color}

\begin{document}\sloppy
\ninept

\def\x{{\mathbf x}}
\def\L{{\cal L}}

\title{Single-View Food Portion Estimation: Learning Image-to-Energy Mappings Using Generative Adversarial Networks} 
%
\name{Shaobo Fang$^{\star}$, Zeman Shao$^{\star}$, Runyu Mao$^{\star}$, Chichen Fu$^{\star}$, 
\thanks{This work was partially sponsored by the National Science Foundation under grant 1657262,  a Healthway Health Promotion Research Grant and from the Department of Health, Western Australia, and by the endowment of the Charles William Harrison Distinguished Professorship at Purdue University. Address all correspondence to Fengqing Zhu, zhu0@ecn.purdue.edu or see www.tadaproject.org.}}
\secondlinename{Deborah A. Kerr$^{\dagger}$, Carol J. Boushey$^{\ddagger}$, Edward J. Delp$^{\star}$ and Fengqing Zhu$^{\star}$}
\address{
$^{\star}$School of Electrical and Computer Engineering, Purdue University, West Lafayette, Indiana, USA\\
$^{\dagger}$School of Public Health, Curtin University, Perth, Western Australia\\
$^{\ddagger}$Cancer Epidemiology Program, University of Hawaii Cancer Center, Honolulu, Hawaii, USA
}

\maketitle
%


\begin{abstract}
Due to the growing concern of chronic diseases and other health problems related to diet, there is a need to develop accurate methods to estimate an individual's food and energy intake. Measuring accurate dietary intake is an open research problem. In particular, accurate food portion estimation is challenging since the process of food preparation and consumption impose large variations on food shapes and appearances. 
In this paper, we present a food portion estimation method to estimate food energy (kilocalories) from food images using Generative Adversarial Networks (GAN).
We introduce the concept of an ``energy distribution'' for each food image.
To train the GAN, we design a food image dataset based on ground truth food labels and segmentation masks for each food image as well as energy information associated with the food image.
Our goal is to learn the mapping of the food image to the food energy.
We can then estimate food energy based on the energy distribution.
We show that an average energy estimation error rate of $10.89\%$ can be obtained by learning the image-to-energy mapping. 
\end{abstract}

\vspace{-0.2cm}
\section{Introduction}
Dietary assessment, the process of determining what someone eats during the course of a day, provides valuable insights for mounting intervention programs for prevention of many chronic diseases.
Traditional dietary assessment techniques, such as dietary record, requires individuals to keep detailed written reports for 3-7 days of all foods or drink consumed~\cite{six2010} and is a time consuming and tedious process.
Smartphones provide a unique mechanism for collecting dietary information and monitoring personal health.
Several mobile dietary assessment systems, that use food images acquired during eating occasions, have been described such as the TADA system~\cite{zhu2010A, zhu-2015}, FoodLog~\cite{foodlogA}, FoodCam~\cite{joutou2009}, DietCam~\cite{kong2012}, and Im2Calories~\cite{Meyers_2015_ICCV} to automatically determine the food types and energy consumed  using image analysis and computer vision techniques. 
Estimating food portion size/energy (kilocalories) is a challenging task since the process of food preparation and consumption impose large variations on food shapes and appearances.  
There are several image based techniques for food portion size estimation that either require a user to take multiple images/videos or modify the mobile device such as the use of multiple images~\cite{Divakaran-2009, kong2012, mougiakakou_2013}, video~\cite{Sun_2010}, 3D range finding~\cite{Shang_2011} and RGB-D images~\cite{Chen:2012aa}.
In this paper we focus on food portion estimation from a single food image since this reduces a user's burden in the number and types of images that need to be acquired~\cite{daugherty2012}.

Estimating food portion size or food volume from a single image is an ill-posed inverse problem.
Most of the 3D information has been lost during the projection process from 3D world coordinates onto 2D camera sensor plane.
Various approaches have been developed to estimate food portion size and energy information from a single-view food image.
In~\cite{Sun_2012}, a 3D model is manually fitted to a 2D food image to estimate the portion size. This approach is not feasible for automatic food portion analysis.
In~\cite{shervin_2015}, food image areas are used for portion size estimation based on user's thumbnail as a size reference.  
In~\cite{divakaran_2015}, the pixels in each corresponding food segment are counted to determine the portion sizes. In~\cite{aizawa_2013}, food image is divided into sub-regions and food portion estimation is done via pre-determined serving size classification.  
We have previously developed a 3D geometric-model based technique for portion estimation~\cite{fang_2015, fang_2017} which incorporates the 3D structure of the eating scene and use geometric models for food objects. 
We showed that more accurate food portion estimates could be obtained using geometric models for food objects whose 3D shape can be well-defined compared to a high resolution RGB-D images~\cite{fang_2016}.
Geometric-model based techniques require accurate food labels and segmentation masks. Errors from these steps can propagate into food portion estimation.

More recently, several groups have developed food portion estimation methods using deep learning~\cite{dlnature} techniques, in particular, Convolutional Neural Networks (CNN)~\cite{lecun_cnn}. 
In~\cite{Meyers_2015_ICCV}, a food portion estimation method is proposed based on the prediction of depth maps~\cite{NYUv2_RGBD} of the eating scene.
However, we have shown that the depth based technique is not guaranteed to produce accurate estimation of food portion~\cite{fang_2016}.
In addition, energy/nutrient estimation accuracy was not reported in ~\cite{Meyers_2015_ICCV}.
In~\cite{yanai_2017}, a multi-task CNN~\cite{multi_task_cnn} architecture was used for simultaneous tasks of energy estimation, food identification, ingredient estimation and cooking direction estimation.
Food calorie estimation is treated as a single value regression task~\cite{yanai_2017} and only one unit in the last fully-connected layer (FC) in the VGG-16~\cite{vgg} is used for calorie estimation.


Although CNN techniques have achieved impressive results for many computer vision tasks, they depend heavily on well-constructed training datasets and proper selection of the CNN architecture.
We propose in this paper to use generative models to estimate the food energy distribution from a single food image.
We construct a food energy distribution image that has a one-to-one pixel correspondence with the food image.
Each pixel in the energy distribution image represents the relative spatial amount (or weight) of food energy at the corresponding pixel location.
Therefore, a food energy distribution image provides insight not only on where the food items are located in the scene, but also reflects the weights of energy in different food regions (for example, regions of the image containing broccoli should have smaller weights due to lower energy (kilocalories) compared to regions of the image containing steak).   
The energy distribution image is one way that we can visualize these relationship. 

More specifically, the generative model is trained on paired images~\cite{pix2pix} mapping a food image to its corresponding energy distribution image.
Our goal is to learn the mapping of the food image to the food energy distribution image so that we can construct an energy distribution image for any eating occasion and then use this energy distribution to estimate portion size.
The weights in food energy distribution image for the training data are assigned based on ground truth energy using a linear transform described in Section~\ref{con-data}. 
We use a Generative Adversarial Networks (GAN) architecture~\cite{gan} as GAN has shown impressive success in training generative models~\cite{pix2pix, HD-cGANs, CycleGAN2017, dc_gans, coupled_gans} in recent years.
Currently, no publicly available food image dataset meets all of the requirements for training our generative model that learns the ``image-to-energy mapping."
We constructed our own dataset based on ground truth food labels, segmentation masks and energy information for training the generative model.
The contribution of this paper is twofold. First, we show that the proposed method can obtain accurate estimates of food energy from a single food image. Second, we introduce a  method for modeling the characteristics of energy distribution in an eating scene. 
\vspace{-0.2cm}      
\section{Learning Image-to-Energy Mappings}
Here we will initially discuss the requirements of the training dataset and then we will describe in more details how we construct the energy distribution image from the training data.
Image pairs consisting of the food image and corresponding energy distribution image are required to train the GAN.
There are a several publicly available food image datasets such as the PFID~\cite{chen2009}, UEC-Food 100/256~\cite{kawano2014automatic} and Food-101~\cite{bossard14}.  
However, none of these dataset contains sufficient information required for training a generative model that we can use to learn the ``image-to-energy mapping".
We created our own paired image dataset for training the GAN with ground truth food labels, segmentation masks and energy/nutrient information from a food image dataset we have collected from dietary studies. This is described in more details in Section~\ref{con-data}.
We use the conditional GAN architecture~\cite{pix2pix} for training our generative model.

\subsection{The Image-to-Energy Dataset}
\label{con-data}
The generative model is designed to best capture the characteristics of the energy distribution associated with food items in an eating scene.
For food types that have different energy distribution (such as broccoli versus steak), the differences should be reflected in the energy distribution image.   
For constructing the image-to-energy training dataset, we use food images collected from a free-living dietary TADA study~\cite{freeliving_11}.
We manually generated the ground truth food label and segmentation mask associated with each food item in the user food image dataset.    
The ground truth energy information (in kilocalories) for each food item was provided by registered dietitians. 
For these food images we have a fiducial marker with known dimension that is located in each eating scene to provide references for worlds coordinates, camera pose, and color calibration.
The fiducial marker is a 5 $\times$ 4 color checkerboard pattern as shown in Figure~\ref{fig:input}.
The food energy distribution image we construct from the above ground truth information needs to reflect the differences in spatial energy distribution for food regions in the scene.
For example, for French fries stacked in pyramid shape, the center region of French fries should have more relative energy weight compared to the edge regions in the energy distribution image.




To construct the energy distribution image we first detect the location of the fiducial marker using~\cite{ZhangCamera}.
We then obtain the 3 $\times$ 3 homography matrix $ \mathbf{H} $ using the Direct Linear Transform (DLT)~\cite{hartley2004} to rectify the image and remove projective distortion.   
Assume $\mathbf{I}$ is the original food image, the rectified image $\hat{\mathbf{I}}$ can then be obtained by: $\hat{\mathbf{I}} = \mathbf{H}^{-1} \mathbf{I}$.
The segmentation mask $S_k$ associated with food $k$ can then be projected from the original pixel coordinates to the rectified image coordinates as $\hat{S}_k = \mathbf{H}^{-1} {S}_k$.
At each pixel location $(\hat{i}, \hat{j}) \in \hat{S}_k$, we assign a  scale factor $\hat{w}_{\hat{i},\hat{j}}$ reflecting the distance of the pixel location $(\hat{i},\hat{j})$ to the centroid of the segmentation mask $\hat{S}_k$.
The scale factor $\hat{w}_{\hat{i},\hat{j}}$ is defined as:
\begin{equation}
\hat{w}_{\hat{i},\hat{j}}  = \frac{1}{\sqrt{(\hat{i} -\hat{i}_c)^2 + (\hat{j}-\hat{j}_c)^2} + \phi_{\hat{S}_k}^{0.5}}, \ \ \ \forall (\hat{i},\hat{j}) \in \hat{S}_k,
\label{eq:weight}
\end{equation}
where $(\hat{i}_c, \hat{j}_c)$ is the centroid of $\hat{S}_k$ and the regularization term, $\phi_{\hat{S}_k}$, is defined as $\phi_{\hat{S}_k} = (\sum\limits_{\forall (\hat{i},\hat{j}) \in \hat{S}_k} \mathbf{1})$.
If the pixel location $(\hat{i},\hat{j})$ is outside of the segmentation mask $\hat{S}_k$, then $\hat{w}_{\hat{i},\hat{j}} = 0, \forall (\hat{i},\hat{j}) \notin \hat{S}_k$. 
With the scale factor $\hat{w}_{\hat{i},\hat{j}}$ assigned to each pixel location in $\hat{S}_k$, we can project the weighted segmentation masks $\hat{S}_k$ back to the original pixel coordinates as $\bar{S}_k = \mathbf{H} \hat{S}_k$, and learn the parameter $\mathbf{\rho}_k$ such that:
\begin{equation}
c_k = \mathbf{\rho}_k \sum\limits_{\forall (\bar{i},\bar{j}) \in \bar{S}_k} \bar{w}_{\bar{i},\bar{j}},
\label{eq:calsum}
\end{equation}  
where $c_k$ is the ground truth energy associated with food $k$, $\mathbf{\rho}_k$ is the energy mapping coefficient for $\bar{S}_k$ and $\bar{w}_{\bar{i},\bar{j}}$ is the energy weight factor at each pixel that makes up the ground truth energy distribution image. 
We then update the energy weight factors in $\bar{S}_k$ as:
\begin{equation}
\bar{w}_{\bar{i},\bar{j}} = \mathbf{\rho}_k \cdot \bar{w}_{\bar{i},\bar{j}}, \ \  \forall (\bar{i},\bar{j}) \in \bar{S}_k.
\label{eq:update}
\end{equation}
We repeat the process following Equation~\ref{eq:weight} and~\ref{eq:calsum} for all $k \in \{1, \dots, M \}$ where $M$ is the number of food items in the eating scene image.
We can then construct a ground truth energy distribution image $\bar{\mathcal{W}}$ of the same size as $\bar{\mathbf{I}}$: $\bar{\mathbf{I}} = \mathbf{H} \hat{\mathbf{I}}$, by overlaying all segments $\bar{S}_k$, $k \in \{1, \dots, M \}$ onto $\bar{\mathcal{W}}$. Thus, we obtain the paired images of an eating scene: the image $\bar{\mathbf{I}}$ and the energy distribution image $\bar{\mathcal{W}}$ with one-to-one pixel correspondence as shown in Figure~\ref{fig:input} and~\ref{fig:ground_truth}. 



\subsection{Learning The Image-to-Energy Mappings}
\label{energy-map}
\begin{figure*}
\subfigure[{Eating occasion image $\bar{\mathbf{I}}$.}]
{
\label{fig:input}
\centering{\epsfig{figure=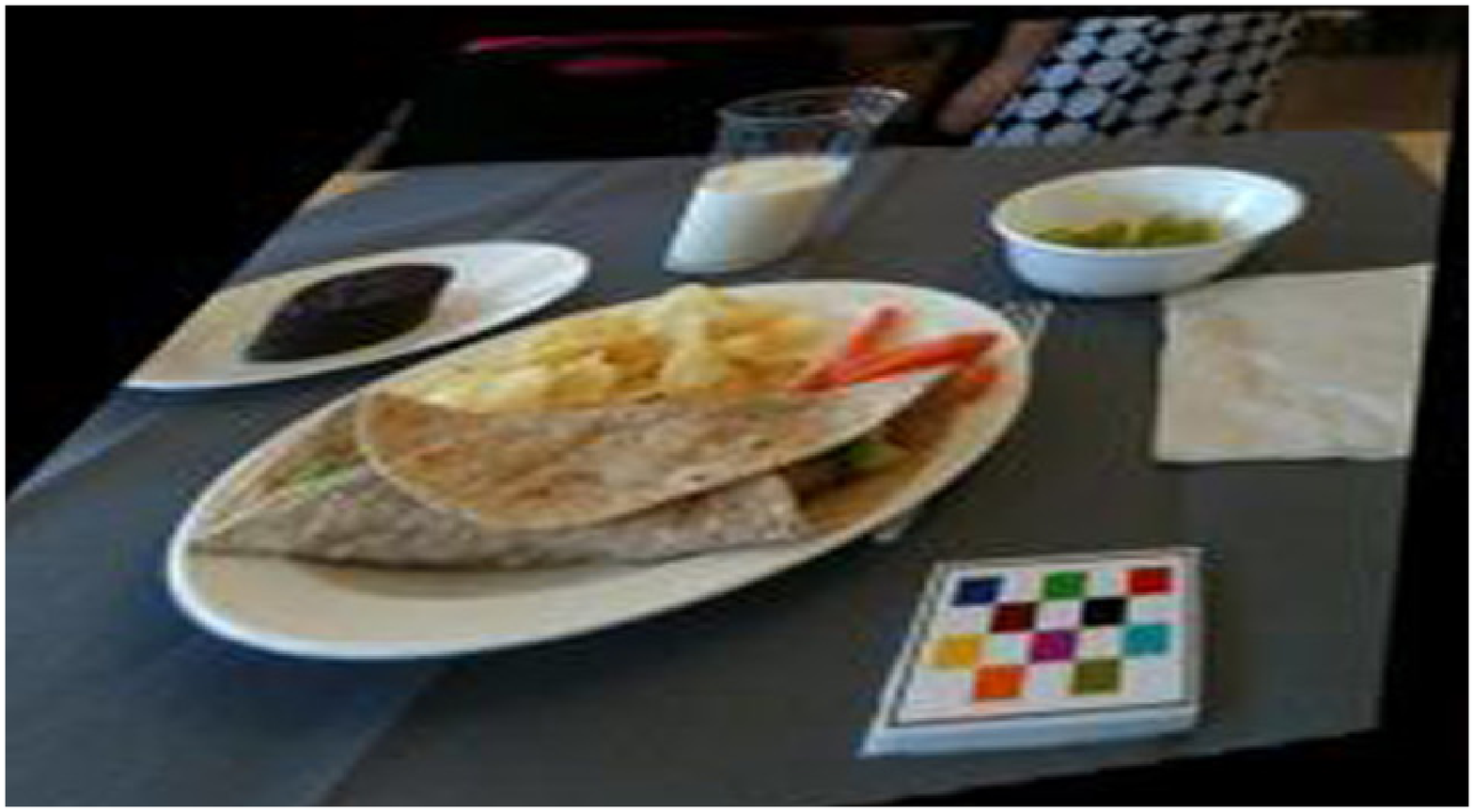, width = 5.6cm}}
}
\hfill
\subfigure[{Ground truth energy distribution image $\bar{\mathcal{W}}$.}]
{
\centering{\epsfig{figure=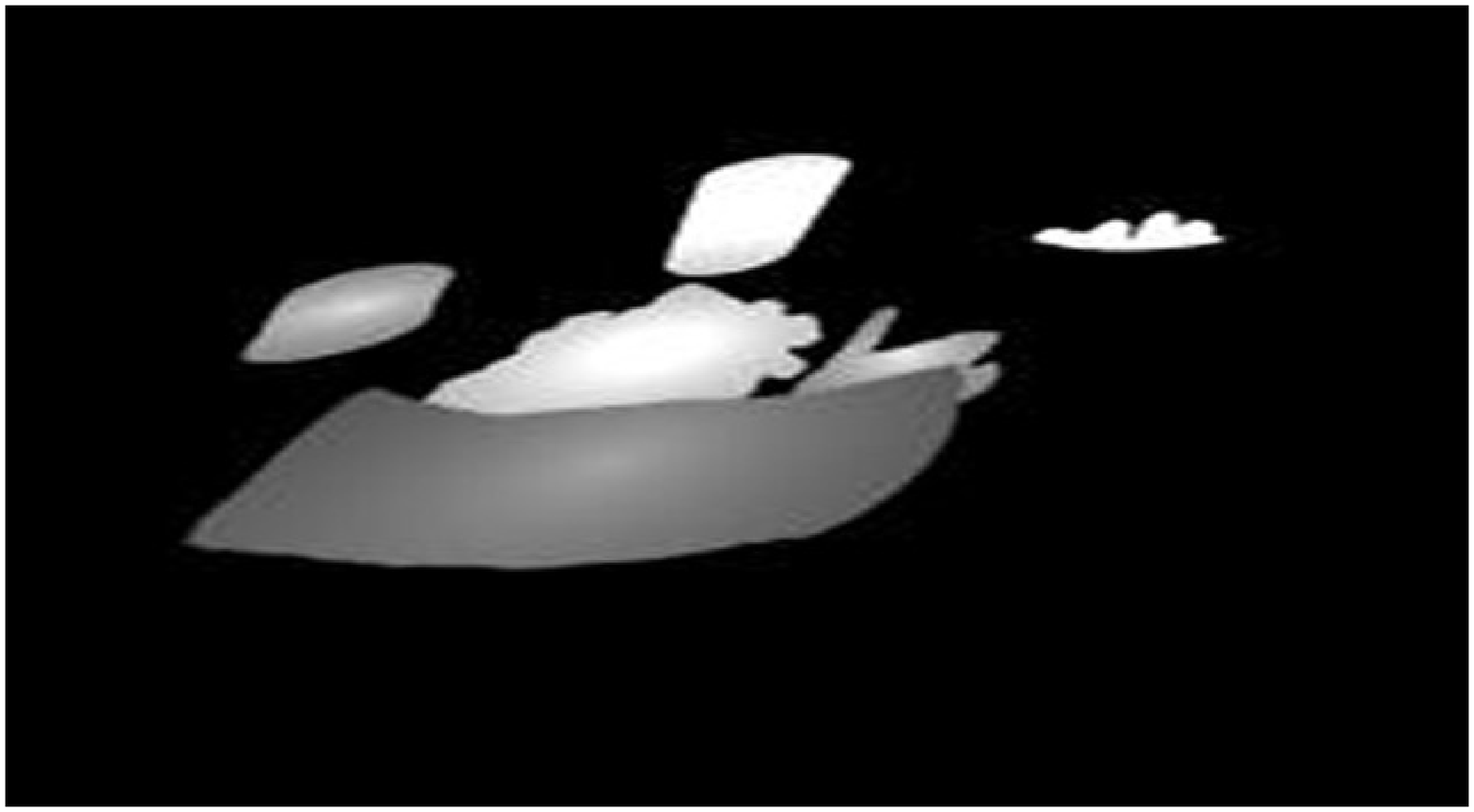, width = 5.6cm}}
\label{fig:ground_truth}
}
\hfill
\subfigure[{Estimated energy distribution image $\tilde{\mathcal{W}}$.}]
{
\centering{\epsfig{figure=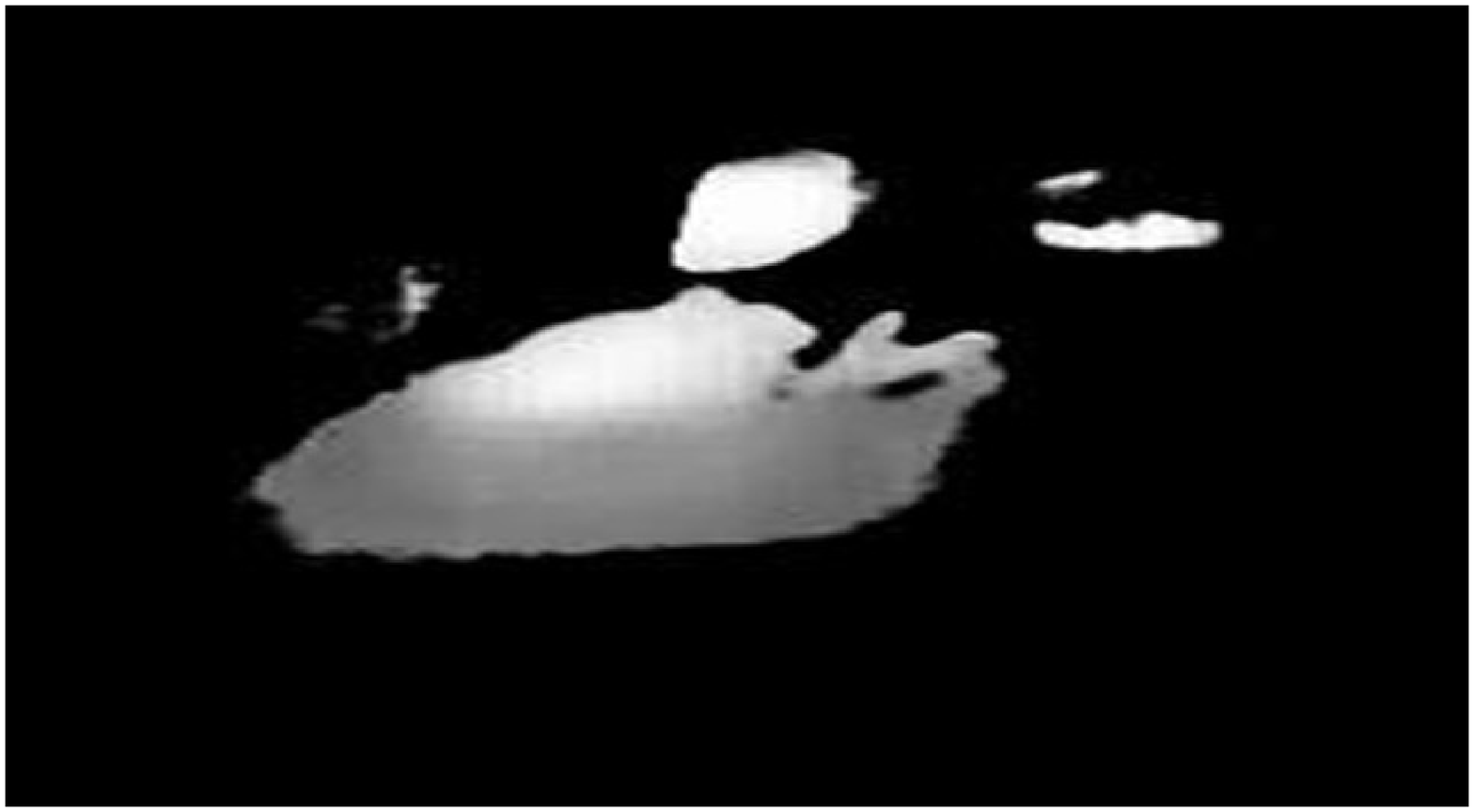, width = 5.6cm}}
\label{fig:output}
}
\label{fig:rgbc_system}
\caption{Learning image-to-energy translation using generative models.}
\end{figure*}


In this paper we use a Conditional GAN (cGAN)~\cite{pix2pix} that learns a generative model under conditional setting based on an input image.
A cGAN is a natural fit for our ``image-to-energy mapping" task since we want to predict the energy distribution image based on a food image.     

More specifically, the  cGAN attempts to learn the mapping from a random noise vector $\mathbf{z}$ to a target image $\mathbf{y}$ conditioned on the observed image $\mathbf{x}$: $G(\mathbf{x}, \mathbf{z}) \rightarrow \mathbf{y}$.
The objective function of a conditional GAN is expressed as:
\begin{equation}
\begin{aligned}
\mathcal{L}_{cGAN}(G, D) &= \mathbb{E}_{\mathbf{x}, \mathbf{y} \sim p_{data}(\mathbf{x}, \mathbf{y})}[\log D(\mathbf{x}, \mathbf{y})] + \\
& \mathbb{E}_{\mathbf{x} \sim p_{data}(\mathbf{x}), \mathbf{z} \sim p_{z}(\mathbf{z})} [\log(1 - D(\mathbf{x}, G(\mathbf{x}, \mathbf{z}))].
\end{aligned}
\label{eq:cGANsobject}
\end{equation}
An additional conditional loss $\mathcal{L}_{conditional}(G)$ is added~\cite{pix2pix} that further improves the generative model's mapping $G(\mathbf{x}, \mathbf{z}) \rightarrow \mathbf{y}$:
\begin{equation}
\mathcal{L}_{conditional}(G) = \mathbb{E}_{\mathbf{x}, \mathbf{y} \sim p_{data}(\mathbf{x}, \mathbf{y}), \mathbf{z} \sim p_{z}(\mathbf{z})}[\mathcal{D}(\mathbf{y}, G(\mathbf{x}, \mathbf{z}))],
\label{eq:closs}
\end{equation}
where $\mathcal{D}(\mathbf{y}, G(\mathbf{x}, \mathbf{z}))$ measure the distance between $\mathbf{y}$ and $G(\mathbf{x}, \mathbf{z})$.
Commonly used criteria for $\mathcal{D}(\cdot)$ are the $L_2$ distance~\cite{efros_2016}:
\begin{equation}
\mathcal{D}(\mathbf{y}, G(\mathbf{x}, \mathbf{z}))  = \frac{1}{n}\sum_{i=1}^{n}(\mathbf{y}_i - G(\mathbf{x}_i, \mathbf{z}_i))^2, 
\label{eq:l2}
\end{equation}
the $L_1$ distance~\cite{pix2pix}: 
\begin{equation}
\mathcal{D}(\mathbf{y}, G(\mathbf{x}, \mathbf{z}))  = \frac{1}{n}\sum_{i=1}^{n}|\mathbf{y}_i - G(\mathbf{x}_i, \mathbf{z}_i)|,
\label{eq:l1}
\end{equation}
and a smooth version of the $L_1$ distance:
\begin{equation}
       \mathcal{D}(\mathbf{y}, G(\mathbf{x}, \mathbf{z})) = \frac{1}{n}\sum_{i=1}^{n} 
        \begin{cases}
            \frac{(\mathbf{y}_i - G(\mathbf{x}_i, \mathbf{z}_i))^2}{2} & \text{if $|\mathbf{y}_i - G(\mathbf{x}_i, \mathbf{z}_i)| < 1$} \\
\ \ \\
            |\mathbf{y}_i - G(\mathbf{x}_i, \mathbf{z}_i)| & \text{otherwise}.
        \end{cases}
\label{eq:smoothl1}
\end{equation}
The final objective for both the cGAN and the conditional terms is defined as~\cite{gan, pix2pix}:
\begin{equation}
G^* = \arg \min_G \max_D \mathcal{L}_{cGAN}(G, D) + \lambda \mathcal{L}_{conditional}(G).
\label{eq:final_loss}
\end{equation}
The generative model $G^*$ obtained from Equation~\ref{eq:final_loss} is then used to predict the energy distribution image $\tilde{\mathcal{W}}$ (Figure~\ref{fig:output}) based on the food image (Figure~\ref{fig:input}). 


\vspace{-0.2cm} 
\section{Experimental Results}
We have 202 food images that have been manually annotated with ground truth segmentation masks and labels as training samples.
All the food images are collected from a free-living (in the wild) TADA dietary study~\cite{freeliving_11}.
Registered dietitians provided the ground truth energy information for each food item in the images. 
We constructed a dataset of paired images based on the 202 food images.
Data augmentation techniques such as rotating, cropping and flipping were used to further expand our training dataset so that a total of 1875 paired images were used to train the cGAN.
We used 220 paired images for testing.

We believe the training dataset size is sufficient for our task of predicting the energy distribution image because the cGAN is a mapping of a higher dimensional food image to a lower dimensional energy distribution image. 
In addition, since all food images are captured by users sitting naturally at a table, there is no drastic changes in viewing angles (for example, from wide angle to close up).
In other image-to-image mapping tasks, a training dataset size of 400 has been used~\cite{pix2pix} for architectural labels (simple features) to photo translation (complex features)~\cite{facade}.


In testing, once the cGAN estimates the energy distribution image $\tilde{\mathcal{W}}$, we can then determine the energy for a food image (portion size estimation) as: estimated energy $= \sum_{\forall (i, j) \in \bar{\mathbf{I}}} (\tilde{w}_{i,j})$.
We compared the estimated energy image $\tilde{\mathcal{W}}$ (Figure~\ref{fig:output}) to the ground truth energy image $\bar{\mathcal{W}}$ (Figure~\ref{fig:ground_truth}), and define the error between $\bar{\mathcal{W}}$ and $\tilde{\mathcal{W}}$ as:
\begin{equation}
\text{Energy Estimation Error Rate} = \frac{\sum_{\forall (i, j) \in \bar{\mathbf{I}}} (\tilde{w}_{i,j} - \bar{w}_{i,j})}{\sum_{\forall (i, j) \in \bar{\mathbf{I}}} (\bar{w}_{i,j})} 
\label{eq:error}
\end{equation}

\begin{figure*}[t]
\hspace{-0.8cm}
\subfigure[{Encoder-decoder.}]
{
\label{fig:encoder_decoder}
\centering{\epsfig{figure=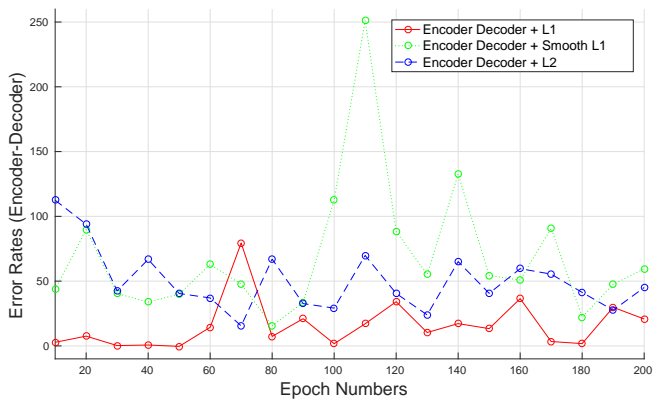, width = 10.1cm}}
}
\hspace{-0.9cm}
\subfigure[{U-Net.}]
{
\centering{\epsfig{figure=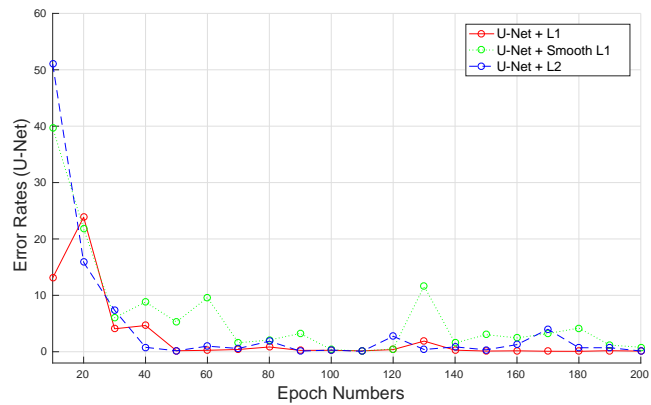, width = 10.1cm}}
\label{fig:u-net}
}
\label{fig:error_rates}
\caption{Comparison of error rates of different generative models: encoder-decoder versus U-Net.}
\end{figure*}

To compare different cGAN models, we used the encoder-decoder architecture~\cite{encoder_decoder} and the U-Net architecture~\cite{unet}.
We compared the energy estimation error rates at different epochs for both architectures.
We observed that the U-Net architecture (Figure~\ref{fig:u-net}) is more accurate in energy estimation and more stable compared to the encoder-decoder architecture (Figure~\ref{fig:encoder_decoder}).
This is due to the fact that the U-Net can copy information from the ``encoder'' layers directly to the ``decoder" layers to provide precise locations~\cite{unet}, an idea similar to ResNet~\cite{resnet}.
 
We also compared the energy estimation error rates under different conditional loss settings: $\mathcal{L}_{conditional}(G)$ using U-Net.
We used the batch size of 16 with $\lambda = 100$ in Equation~\ref{eq:final_loss},  the Adam~\cite{adam} solver with initial learning rate $\alpha = 0.0002$, and momentum parameters $\beta_1 = 0.5$, $\beta_2 = 0.999$ as in~\cite{pix2pix}. 
Based on our experiments, distance measure $\mathcal{D}(\cdot)$ using the $L_1$ or $L_2$ norms is better than using smoothed $L_1$ norm.
At epoch 200, the energy estimation error rates are $10.89\%$ (using $L_1$ criterion) and $12.67\%$ (using $L_2$ criterion), respectively.
Using geometric-models~\cite{fang_2015} techniques, the energy estimation error was $35.58\%$ ~\cite{fang_2017}.

\vspace{-0.2cm}
\section{Conclusion}
In this paper, we presented a food portion estimation technique based on generative models.
We showed that the energy estimation error was reduced to  $10.89\%$.
Compared to earlier geometric model-based technique that relies on accurate food segmentation and classification, the energy estimation task can be performed in parallel with segmentation and classification.
We are interested in incorporating the results from different parts of our system (classification, segmentation and energy estimation) to further improve the overall accuracy of our dietary assessment system.
We are also investigating the effect of different eating scene characteristics on the energy estimation error. For example, food items that are occluded often yield high energy estimation error. 
\vspace{-0.2cm}
\ninept
\bibliographystyle{IEEEbib}
\bibliography{icip2018}

\end{document}